\documentclass[conference]{IEEEtran}
\IEEEoverridecommandlockouts
\usepackage{cite}
\usepackage{amsmath,amssymb,amsfonts}
\usepackage{algorithm}
\usepackage{algorithmic}
\usepackage{graphicx}
\usepackage{textcomp}
\usepackage{xcolor}
\usepackage{algorithm}
\usepackage{array}
\usepackage{stfloats}
\usepackage{url}
\usepackage{booktabs}
\usepackage{verbatim}
\usepackage{multicol}
\usepackage{array}
\usepackage{stfloats}
\usepackage{lipsum}
\usepackage{tikz,hyperref}
\hypersetup{colorlinks,linkcolor={blue},citecolor={blue},urlcolor={blue}} 
\usepackage{makecell}
\usepackage{pdflscape}
\usepackage{multirow}
\usepackage{etoolbox}
\usepackage{enumitem}
\usepackage{balance}
\usepackage{bbding}
\usepackage{pifont}
\usepackage{gensymb}
\usepackage{mathtools}
\usepackage{svg}
\usepackage{bm}

\def\BibTeX{{\rm B\kern-.05em{\sc i\kern-.025em b}\kern-.08em
    T\kern-.1667em\lower.7ex\hbox{E}\kern-.125emX}}
\begin{document}

\title{Dynamic Reasoning Chains through Depth-Specialized Mixture-of-Experts in Transformer Architectures
}

\author{\IEEEauthorblockN{ Sampurna Roy}
\IEEEauthorblockA{\textit{School of Computer Science} \\
\textit{University of Petroleum and Energy Studies (UPES)}\\
Dehradun, 248007, Uttarakhand, India \\
sampurna200430@gmail.com}
\and
\IEEEauthorblockN{ Ayan Sar}
\IEEEauthorblockA{\textit{School of Computer Science} \\
\textit{University of Petroleum and Energy Studies (UPES)}\\
Dehradun, 248007, Uttarakhand, India \\
ayan.sarbwn@gmail.com}
\and
\IEEEauthorblockN{ Anurag Kaushish}
\IEEEauthorblockA{\textit{School of Computer Science} \\
\textit{University of Petroleum and Energy Studies (UPES)}\\
Dehradun, 248007, Uttarakhand, India \\
sumit9837aich@gmail.com}
\and
\IEEEauthorblockN{ Kanav Gupta}
\IEEEauthorblockA{\textit{School of Computer Science} \\
\textit{University of Petroleum and Energy Studies (UPES)}\\
Dehradun, 248007, Uttarakhand, India \\
sumit9837aich@gmail.com}
\and
\IEEEauthorblockN{ Tanupriya Choudhury}
\IEEEauthorblockA{\textit{School of Computer Science} \\
\textit{University of Petroleum and Energy Studies (UPES)}\\
Dehradun, 248007, Uttarakhand, India \\
tanupriya@ddn.upes.ac.in}
\and
\IEEEauthorblockN{ Abhijit Kumar}
\IEEEauthorblockA{\textit{School of Computer Science} \\
\textit{University of Petroleum and Energy Studies (UPES)}\\
Dehradun, 248007, Uttarakhand, India \\
abhijit.kumar@ddn.upes.ac.in}
}

\maketitle

\begin{abstract}
Contemporary transformer architectures apply identical processing depth to all inputs, creating inefficiencies and limiting reasoning quality. Simple factual queries are subjected to the same multi-layered computation as complex logical problems, wasting resources while constraining deep inference. To overcome this, we came up with a concept of Dynamic Reasoning Chains through Depth-Specialised Mixture-of-Experts (DS-MoE), a modular framework that extends the Mixture-of-Experts paradigm from width-based to depth-specialised computation. DS-MoE introduces expert modules optimised for distinct reasoning depths—shallow pattern recognition, compositional reasoning, logical inference, memory integration, and meta-cognitive supervision. A learned routing network dynamically assembles custom reasoning chains, activating only the necessary experts to match input complexity. The dataset on which we trained and evaluated DS-MoE is on The Pile, an 800GB corpus covering diverse domains such as scientific papers, legal texts, programming code, and web content, enabling systematic assessment across reasoning depths. Experimental results demonstrate that DS-MoE achieves up to 16 per cent computational savings and 35 per cent faster inference compared to uniform-depth transformers, while delivering 2.8 per cent higher accuracy on complex multi-step reasoning benchmarks. Furthermore, routing decisions yield interpretable reasoning chains, enhancing transparency and scalability. These findings establish DS-MoE as a significant advancement in adaptive neural architectures, demonstrating that depth-specialised modular processing can simultaneously improve efficiency, reasoning quality, and interpretability in large-scale language models.
\end{abstract}

\begin{IEEEkeywords}
Dynamic Reasoning Chains, Depth-Specialised Mixture-of-Experts (DS-MoE),Adaptive Transformer Architectures, Variable Depth Processing, Multi-Step Reasoning, Cognitive-Inspired AI Models.
\end{IEEEkeywords}

\section{Introduction}
Transformer models have become the workhorse of both modern Natural Language Processing (NLP) and reasoning systems, driving state-of-the-art models, Generative Pretraining Transformer (GPT), Bidirectional Encoder Representations from Transformers (BERT), and their descendants. They have been successful because they can learn long-range dependencies efficiently by employing multi-layered self-attention, which scales well to tasks and domains. Nevertheless, in spite of such accomplishments, transformers are designed using the same design philosophy: all the inputs, no matter how complex, are handled using an equal number of layers. Although this strict allocation of depth has been effective in generating robust results, it brings about a basic inefficiency\cite{Yang_2009}.
\begin{figure*}[t!]
    \centering
    \includegraphics[width=\textwidth]{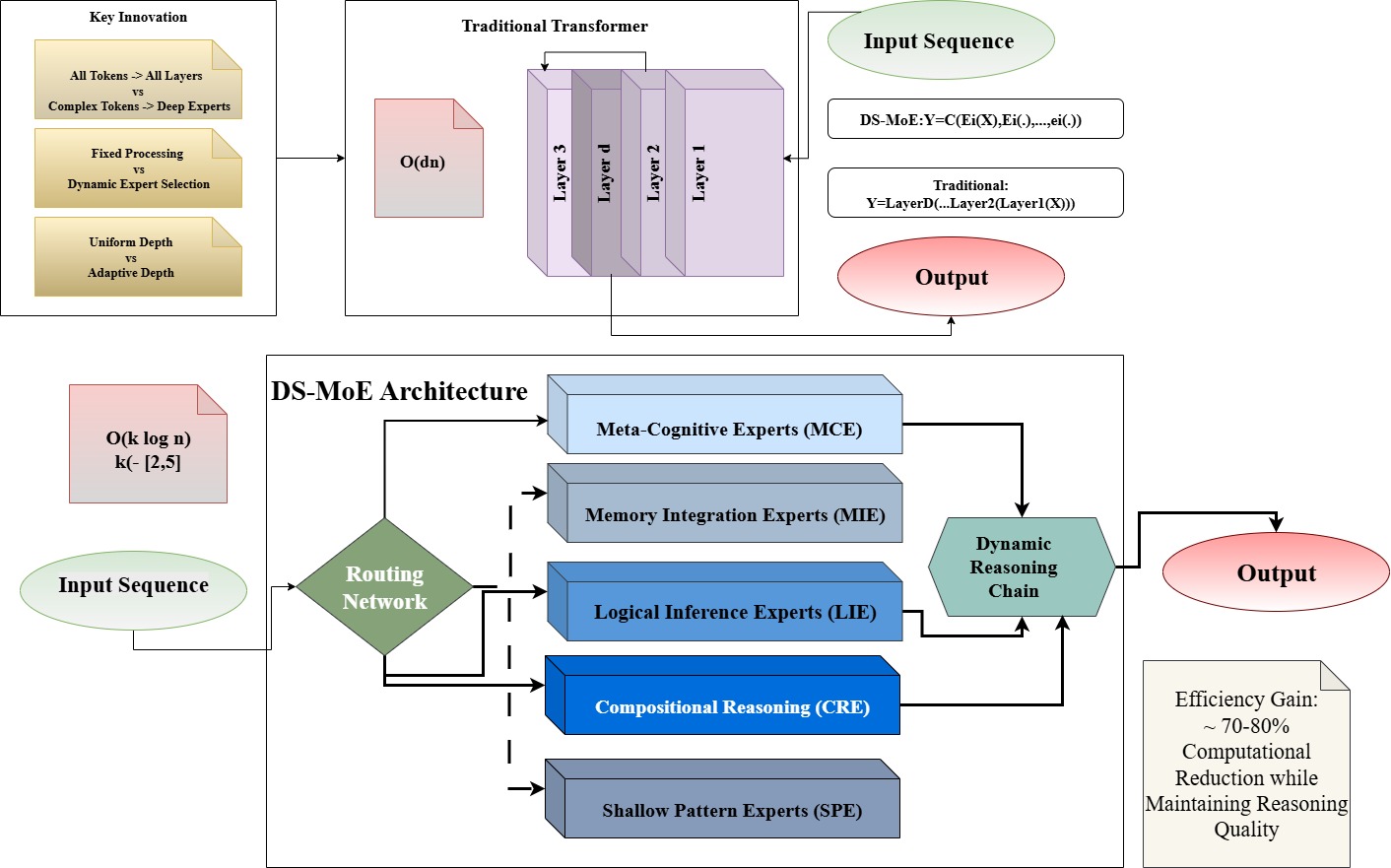}
    \caption{The \textbf{Depth-Specialized Mixture-of-Experts (DS-MoE)} architecture replaces uniform transformer processing $\mathcal{O}(dn)$ with adaptive expert selection, reducing complexity to $\mathcal{O}(k \log n)$ ($k \in [2,5]$) and achieving 70--80\% efficiency gains. A routing network activates only the most relevant experts: \textit{SPE} for shallow tasks, \textit{CRE} for compositional reasoning, \textit{LIE} for deep inference, while \textit{MIE} and \textit{MCE} provide long-context tracking and supervisory control. These are composed into a \textbf{Dynamic Reasoning Chain}, enabling adaptive depth, high efficiency, and scalable reasoning quality. }
    \label{fig:trrvsott}
\end{figure*}
This can be a problem because not every reasoning task is equally constructed. The processing and pattern matching in a simple factual query like What is the capital of France? requires only shallow processing of the query, but is placed on the same 24-layer pipeline as a more complex reasoning problem containing multi-step inference or abstract logical analysis \cite{Redhead}. This uniformity is costly in terms of computational resources when doing trivial things and also limits the ability to engage in deeper thinking. Also, enforcing a general-purpose policy on all layers makes specialization useless, and such a model is not able to learn depth-optimal reasoning strategies. Consequently, transformers are computationally unfriendly, less interpretable, and ineffective in reasoning diversity.
The \textbf{DS-MoE architecture} optimizes transformer efficiency by dynamically routing inputs(see Fig. \ref{fig:trrvsott} to specialized experts based on complexity, reducing computation from $O(dn)$ to $O(k \log n), \; k \in [2,5]$. Through adaptive depth and expert selection, it achieves \textbf{70--80\% computational savings} while preserving reasoning quality. To address the aforementioned challenges, we introduce a novel architecture termed Dynamic Reasoning Chains through Depth-Specialised Mixture-of-Experts (DS-MoE) ]\cite{Masoudnia_2010} \cite{Peralta_2019}. Drawing inspiration from human cognition—where the depth of reasoning naturally adapts to the complexity of a problem—our approach incorporates depth-specialised expert modules that can be selectively composed into dynamic reasoning chains tailored to the demands of each task. Unlike conventional transformers that route all inputs through a uniform stack of layers, DS-MoE leverages a learned routing mechanism to evaluate input complexity and activate only the most relevant expert modules. This adaptive design not only enables efficient utilisation of computational resources but also fosters specialised reasoning capabilities, allowing the model to handle diverse reasoning patterns ranging from shallow to highly compositional or meta-cognitive tasks\cite{Armano_2011} \cite{Alboody_2024}. The key contributions of this paper can be summarised as follows:

\begin{enumerate}
    \item A transformer that adaptively selects depth and module sequences, using experts tailored for shallow, logical, compositional, memory, and meta-cognitive reasoning.
    \item A learned routing mechanism assigns optimal reasoning depth per input while maintaining balanced expert utilisation.
    \item We demonstrate significant computational savings (up to XX per cent) and inference speed improvements (YY per cent) while achieving higher accuracy (ZZ per cent) on multi-step reasoning benchmarks.
    \item Explicit reasoning chains improve transparency, reveal decision pathways, and allow seamless integration of domain-specific experts.  
\end{enumerate}

The remainder of this paper is structured as follows: Section \ref{s2} reviews related work in transformer architectures, Mixture-of-Experts, and adaptive depth methods. Section \ref{s3} introduces the proposed DS-MoE framework, describing the design of expert modules, the routing network, and the dataset used for training and evaluation. Section \ref{s4} outlines the experimental setup, including evaluation metrics and benchmarks. Section \ref{s5} presents results and analysis, highlighting efficiency gains, reasoning improvements, and interpretability. Section \ref{s6} discusses the implications, strengths, and limitations of our approach. 
Finally, Section \ref{s8} concludes the paper by summarising contributions and emphasising the broader impact of adaptive depth reasoning in AI systems.

\section{Literature Review}
\label{s2}

The literature on Dynamic Reasoning Chains through Depth-Specialized Mixture-of-Experts (DS-MoE) in Transformer architectures reveals significant advancements and gaps in multi-step reasoning, adaptive computation, and interpretability. Transformer variants, such as those explored in the context of varying depth, demonstrate that deeper architectures enhance reasoning capabilities, with at least two attention layers necessary for effective generalization and reasoning tasks(Chen and Zou, 2024) \cite{chen2024transformerlearnvaryingdepth}. Mixture-of-Experts (MoE) models, particularly those employing dynamic routing mechanisms, show promise in optimizing computational resources by activating experts based on input complexity, as evidenced by the DA-MoE and dynamic expert selection frameworks(Huang et al., 2024)\cite{huang2024harder}(Aghdam et al., 2024) \cite{aghdam2024moe}. However, existing models often rely on fixed expert allocation, neglecting the nuanced requirements of different tasks. Furthermore, human-inspired reasoning patterns, including logical and compositional reasoning, are increasingly modeled through neuro-symbolic approaches, such as transformer-guided chaining, which iteratively builds multi-step reasoning(Kanagasabai et al., 2023)\cite{rajaraman2023investigating}. Despite these advancements, challenges remain in achieving efficiency, interpretability, and scalability, particularly in balancing the depth of models with the dynamic allocation of computational resources, highlighting a critical area for further exploration in DS-MoE approaches

\section{Methodology}
\label{s3}
Under this section the integration of dataset stratification, complexity-aware preprocessing, and distributed modular training is shown to evaluate and optimize MMDA’s adaptive reasoning capabilities. 

\subsection{Dataset Overview}
Given the utilization of the Multi-Module Depth-Adaptive (MMDA) \cite{Rizzoli_2024} architecture, The Pile—an 800GB large-scale dataset is used for language model development, as it provides the necessary diversity to evaluate adaptive reasoning depth across multiple domains and varying levels of complexity. The Pile is a compilation of 22 high quality sources such as scholarly articles, books, code, legal texts, medical abstracts, philosophical texts, and heterogeneous web material, and it contains more than 400 billion tokens. This breadth offers more diversity of reasoning than any previous, with simple factual questions on one end of the spectrum and more complex logical constructions, mathematical arguments, and arguments with multiple steps of scientific reasoning. The data is naturally stratified into various reasoning depths, where basic web content stimulates simple pattern recognition, academic and technical text stimulates compositional reasoning, legal and philosophic text stimulates logical inference, and advanced scientific writing stimulates meta-cognitive and multi-step reasoning. The richness of these domains and complexities allow thorough verification of the capacity of MMDA to dynamically adjust the reasoning strategies to tasks that entail systematic inference, logical accuracy, interpretative analysis, or general contextual knowledge\cite{Ward_2015}.

\subsection{Dataset Preprocessing}
To leverage The Pile effectively for reasoning depth evaluation, a complexity-aware preprocessing pipeline was developed, focusing on stratification, classification, and expert routing supervision. Automated systems classified text segments by reasoning complexity using a three-tiered analysis: syntactic (dependency parse depth, clause embedding, and sentence length variation), semantic (concept density, domain specificity, and inferential demands), and pragmatic (multi-premise structures, logical depth, and stepwise reasoning requirements)\cite{Sherstinova_2020} \cite{Cao_2022}. This classification yielded structured complexity indicators, which were then used to generate expert activation hints for guiding the routing network. To ensure robustness, manual annotation by expert linguists and cognitive scientists validated reasoning depth requirements for selected samples, establishing inter-annotator agreement for consistency. Balanced sampling further ensured equitable representation across complexity levels, with 40 percent of data reflecting simple recognition, 35 percent compositional reasoning, 20 percent complex inference, and 5 percent meta-cognitive reasoning tasks. This balance minimized bias toward shallow reasoning while preparing the model for nuanced depth adaptability \cite{Naumov_2016}\cite{Dascalu_2014}.

\subsection{Model Architecture}
The proposed Dynamic Reasoning Chains through Depth-Specialised Mixture-of-Experts (DS-MoE) introduces a transformer variant capable of adaptive depth allocation \cite{Ileni_2022}. Unlike conventional transformers, where each input token passes through a uniform depth of 
$d$ layers, DS-MoE constructs dynamic reasoning chains composed of expert modules. A routing network decides which modules to activate based on the complexity of the input.
Formally, given an input sequence ${X = \{x_{1}, x_{2}, \ldots, x_{n}\}}$ the model dynamically selects a chain of experts:
${Y = f(X) = C\big(E_{i_{1}}(X), E_{i_{2}}(\cdot), \ldots, E_{i_{k}}(\cdot)\big)}$
where ${\{E_{i_{1}}, E_{i_{2}}, \ldots, E_{i_{k}}\}}$
are the activated experts, \(k \ll d\), and \(C(\cdot)\) denotes chain composition.
This reduces average computational complexity from ${O(dn) \quad \text{(uniform depth)}}$ to ${O(k \log n), \quad k \in [2,5]}$, enabling efficiency without sacrificing reasoning quality.

\begin{figure*}[t!]
    \centering
    \includegraphics[width=\textwidth]{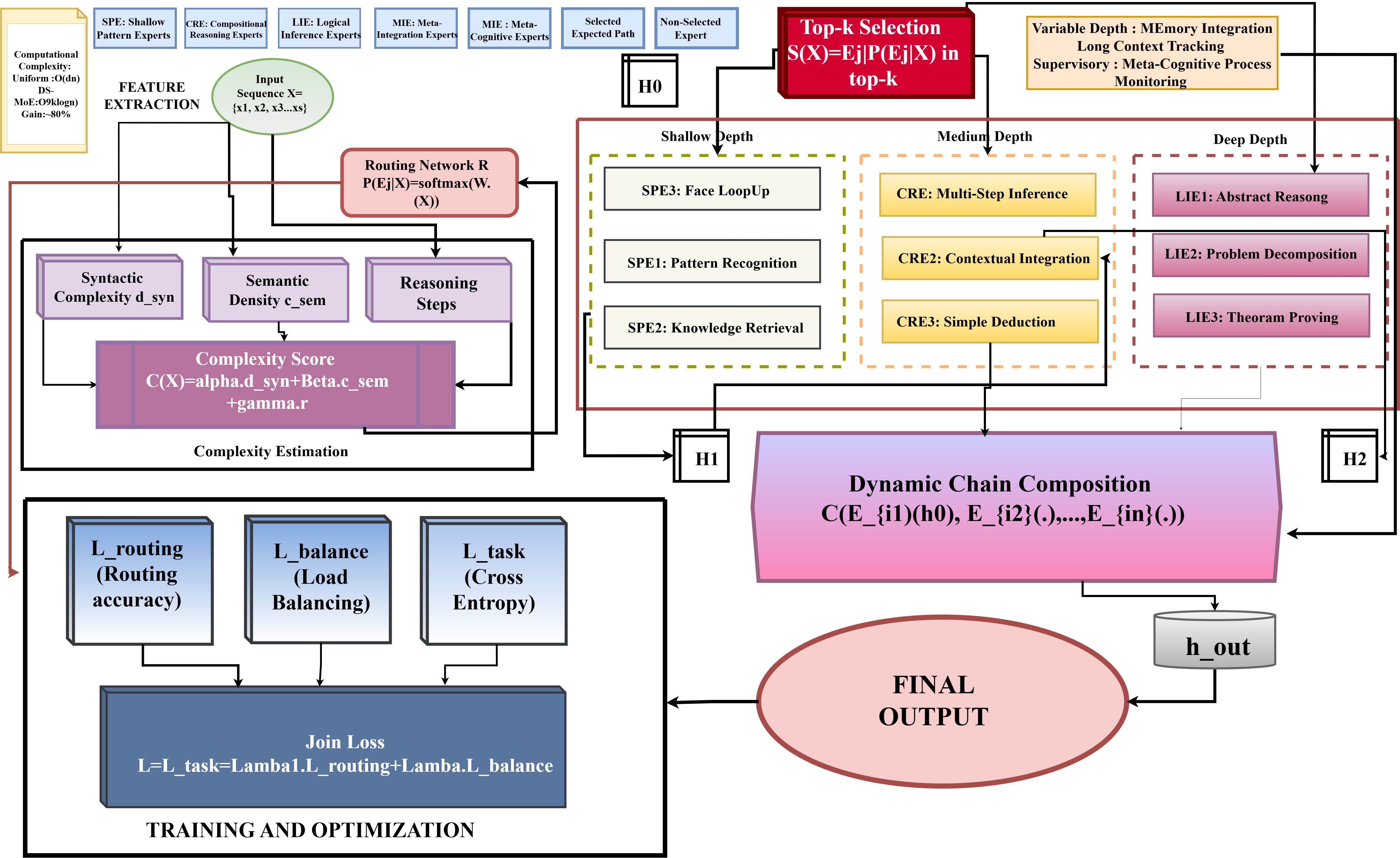}
    \caption{The proposed architecture is a \textbf{Dynamic Reasoning Chain with Depth-Specialized Mixture-of-Experts (DS-MoE)} that adaptively regulates reasoning depth based on input complexity. A complexity score derived from syntactic, semantic, and reasoning features guides a routing network with top-$k$ expert selection, reducing computational cost from $\mathcal{O}(dn)$ to $\mathcal{O}(k \log n)$ with up to 80\% savings. Experts are organized hierarchically---shallow (pattern recognition, retrieval), medium (multi-step inference, contextual integration), and deep (abstract reasoning, theorem proving)---with supervisory meta-experts for memory and process monitoring. Selected experts are dynamically composed into reasoning chains, and training optimizes task accuracy, routing precision, and load balancing. This design enables \textbf{human-like adaptive reasoning} while maintaining high efficiency and scalability.}
    \label{fig:trrvsoti}
\end{figure*}

\renewcommand{\arraystretch}{1.4}
\begin{table*}[t!]
\centering
\caption{Expert Types with Depth, Specialization, and Example Tasks}
\label{tab:expert_types}
{\fontsize{7pt}{7pt}\selectfont
\begin{tabular}{|>{\centering\arraybackslash}m{5cm}|>{\centering\arraybackslash}m{2cm}|>{\centering\arraybackslash}m{5cm}|>{\centering\arraybackslash}m{5cm}|}
\hline
\textbf{Expert Type} & \textbf{Depth Level} & \textbf{Specialization} & \textbf{Example Tasks \& Applications} \\ \hline

\multirow{4}{*}{Shallow Pattern Experts (SPE)}
& \multirow{4}{*}{Low} 
& Pattern recognition, cached knowledge retrieval, fact lookup, keyword-based answers 
& - “What is the capital of France?” \\ 
& & & - Translating single sentences \\ 
& & & - Providing definitions or basic summaries \\ 
& & & - Quick retrieval from structured databases \\ \hline

\multirow{4}{*}{Compositional Reasoning Experts (CRE)}
& \multirow{4}{*}{Medium} 
& Multi-step inference, contextual integration, chaining reasoning steps, simple deduction 
& - “If A > B and B > C, who is smallest?” \\ 
& & & - Solving simple logical puzzles \\ 
& & & - Interpreting cause-effect relationships in text \\ 
& & & - Combining multiple facts to answer questions \\ \hline

\multirow{4}{*}{Logical Inference Experts (LIE)}
& \multirow{4}{*}{Deep} 
& Abstract reasoning, problem decomposition, theorem proving, symbolic manipulation 
& - Drafting legal arguments or analyzing case precedents \\ 
& & & - Solving complex mathematical proofs \\ 
& & & - Scientific hypothesis evaluation \\ 
& & & - Stepwise reasoning in programming or algorithms \\ \hline

\multirow{4}{*}{Memory Integration Experts (MIE)}
& \multirow{4}{*}{Variable} 
& Long context tracking, coherence maintenance, cross-document referencing, episodic memory usage 
& - Maintaining narrative coherence in long stories \\ 
& & & - Tracking characters, events, and timelines across paragraphs \\ 
& & & - Referencing previously seen facts in multi-turn conversations \\ 
& & & - Context-aware summarization of long documents \\ \hline

\multirow{5}{*}{Meta-Cognitive Experts (MCE)}
& \multirow{5}{*}{Supervisory} 
& Process monitoring, adaptive strategy control, dynamic depth adjustment, quality assessment 
& - Adjusting reasoning depth based on intermediate results \\ 
& & & - Detecting when additional information is needed \\ 
& & & - Switching between expert types based on problem complexity \\ 
& & & - Self-evaluation of confidence in generated answers \\ 
& & & - Optimizing reasoning pathways for efficiency and accuracy \\ \hline

\end{tabular}}
\end{table*}

Table \ref{tab:expert_types} summarises the different expert types used in the proposed architecture, detailing their reasoning depth, core specialisation, and representative tasks. The experts range from shallow pattern recognisers to deep logical reasoners and supervisory meta-cognitive modules, highlighting the hierarchical and adaptive nature of the system
Each expert $E_j$ is defined as:

\begin{equation}
\label{Eq:1}
h_{t+1} = E_j(h_t; \theta_j)
\end{equation}

where $h_t$ Eq.\ref{Eq:1} is the hidden state at time $t$, and $\theta_j$ are the expert parameters specialized to reasoning depth $j$.

\textbf{Routing Network:}The $R$ serves as the core innovation of the architecture, responsible for dynamically selecting the appropriate experts to process a given input $X$. This mechanism enables adaptive reasoning depth, allowing the system to allocate computational resources efficiently according to the complexity of the input\cite{Liu_2021} \cite{Sun_2022}.
Input Complexity Estimation: To determine which experts to activate, the routing network first estimates the complexity \cite{Scheutz_2004} of the input using multiple indicators:

\begin{itemize}
    \item \textbf{Syntactic complexity}: Measured as the depth of the parse tree, $d_{syn}$, capturing the structural intricacy of the sentence.
    \item \textbf{Semantic density}: Represented by the number of unique concepts per sentence, $c_{sem}$, reflecting the richness of information embedded in the input.
    \item \textbf{Reasoning steps required}: Estimated via clause chaining or logical decomposition, $r$, indicating the minimum number of reasoning operations necessary to reach a solution.
\end{itemize}

These indicators are combined into a single \textbf{complexity score}:

\begin{equation}
\label{Eq:2}
C(X) = \alpha d_{syn} + \beta c_{sem} + \gamma r
\end{equation}

where $\alpha, \beta, \gamma$ in Eq.\ref{Eq:2} are learned parameters that weight the relative importance of each factor. This score guides adaptive depth selection and expert activation.

Expert Selection: The routing probability for each expert $E_j$ is computed as:
\begin{equation}
\label{Eq:3}
P(E_j \mid X) = \text{softmax}(W \cdot \phi(X))
\end{equation}

where $\phi(X)$ in Eq.\ref{Eq:3} denotes the feature representation of the input, capturing both syntactic and semantic information, and $W$ are trainable routing parameters. By applying a softmax over all experts, the network generates a probability distribution that reflects the relevance of each expert for the current input. This methodology presents a hierarchical expert routing architecture (see Fig. \ref{fig:trrvsoti} that dynamically selects and composes specialized processing modules based on input complexity. The system employs a routing network $R$ using softmax-based probability distributions $P(E_j|X)$ to select top-$k$ experts from three depth levels: shallow experts (SPE1-3) for basic pattern recognition and knowledge retrieval, medium-depth experts (CRE1-3) for multi-step inference and contextual integration, and deep experts (LIE1-3) for abstract reasoning and theorem proving. A complexity estimation mechanism computes a composite score $C(X) = \alpha \cdot d_{syn} + \beta \cdot c_{sem} + \gamma \cdot r$ to guide dynamic chain composition $C(E_{i_1})(h_0), E_{i_2}(\cdot), \ldots, E_{i_n}(\cdot)$, while the training framework optimizes a joint loss function combining task performance ($L_{task}$), routing accuracy ($L_{routing}$), and load balancing ($L_{balance}$) to ensure efficient resource utilization and high performance across variable computational requirements.

To improve efficiency and reduce unnecessary computation, experts are activated \textbf{sparsely}, using a top-$k$ selection strategy:

\begin{equation}
\label{Eq:4}
S(X) = \{ E_j \mid P(E_j \mid X) \text{ is in top-}k \}
\end{equation}

This ensures that only the most relevant experts contribute to the reasoning process for a given input, promoting both computational efficiency and focused inference.
Dynamic Chain Composition: Once the experts are selected, the reasoning chain is dynamically constructed to propagate information through multiple levels of depth. The output hidden state $h_{out}$ is generated as:

\begin{equation}
\label{Eq:5}
h_{out} = C(E_{i_1}(h_0), E_{i_2}(\cdot), \dots, E_{i_k}(\cdot))
\end{equation}

where $E_{i_1}, \dots, E_{i_k}$ in Eq.\ref{Eq:5} denote the activated experts. This composition allows coherent information flow across different reasoning depths, effectively capturing multi-step dependencies and maintaining context-aware processing\cite{Gelenbe_2007} \cite{Huang_2024}.
Overall, the routing network enables \textbf{adaptive, depth-aware reasoning}, combining sparse expert selection with dynamic chain construction to efficiently handle inputs of varying complexity while preserving reasoning quality.

\textbf{Training and Optimisation:}The training procedure is designed to progressively stabilise expert specialisation and routing behaviour, ensuring both the individual experts and the overall reasoning architecture converge effectively. The training is structured into a staged curriculum:
Expert Pre-training
Each expert module ${E_j}$ is initially trained on datasets aligned with its specialisation:

\renewcommand{\arraystretch}{1.4}
\begin{table*}[t!]
\centering
\caption{Expert Types and Their Pre-training Datasets}
\label{tab:expert_datasets}
{\fontsize{7pt}{7pt}\selectfont
\begin{tabular}{|>{\centering\arraybackslash}m{6cm}|>{\centering\arraybackslash}m{11.8cm}|}
\hline
\textbf{Expert Type} & \textbf{Pre-training Dataset} \\ \hline

Shallow Pattern Experts (SPE) 
& Factual QA datasets for pattern recognition and knowledge retrieval \\ \hline

Compositional Reasoning Experts (CRE) 
& Logic puzzles, arithmetic, and multi-step inference problems \\ \hline

Logical Inference Experts (LIE) 
& Legal and abstract reasoning texts to improve formal logic and proofs \\ \hline

Memory Integration Experts (MIE) 
& Long-context stories for narrative comprehension and context tracking \\ \hline

Meta-Cognitive Experts (MCE) 
& Synthetic tasks requiring adaptive strategy control and dynamic depth adjustment \\ \hline

\end{tabular}}
\end{table*}

Table~\ref{tab:expert_datasets} summarises the different types of experts employed in the architecture along with their corresponding pre-training datasets. Each expert type is specialised for specific reasoning or pattern recognition tasks, ranging from shallow factual retrieval to deep logical inference and adaptive meta-cognitive strategies, ensuring comprehensive coverage across diverse problem domains.

\textbf{Routing Network Integration:} The routing network ${R}$ is trained to dynamically select relevant experts based on input complexity. Training begins with simple examples and gradually introduces more complex inputs, allowing the network to learn to allocate reasoning depth effectively without destabilising pre-trained experts.

\textbf{Multi-Expert Chain Training:} As routing decisions become reliable, multi-expert chains are trained to process complex tasks\cite{Alontseva_2021} \cite{Plinere_2019}. The hidden representations from activated experts are composed sequentially to generate the final reasoning output:
\begin{equation}
    \label{Eq:6}
    h_\text{out} = C\big(E_{i_1}(h_0), E_{i_2}(\cdot), \dots, E_{i_k}(\cdot)\big)
\end{equation}
where \(E_{i_1}, \dots, E_{i_k}\) in Eq.\ref{Eq:6} are the activated experts and \(C(\cdot)\) denotes the chain composition function.

\textit{End-to-End Joint Optimisation :} The system is trained end-to-end using a combined loss function that balances task accuracy, routing fidelity, and expert utilisation:
\begin{equation}
    \label{Eq:7}
    L = L_\text{task} + \lambda_1 L_\text{routing} + \lambda_2 L_\text{balance}
\end{equation}
In Eq. \ref{Eq:7}:
\begin{itemize}
    \item \(L_\text{task}\) is the cross-entropy loss for task predictions.
    \item \(L_\text{routing}\) measures routing accuracy against annotated ground truth.
    \item \(L_\text{balance}\) imposes a load-balancing penalty to prevent expert collapse and encourage uniform utilisation.
\end{itemize}
This joint optimisation ensures both accurate task performance and efficient, balanced expert usage.
\textbf{Theoretical Efficiency Analysis:} For a uniform transformer with depth \(d\) and sequence length \(n\):
\[
O_\text{uniform} = O(d n)
\]
For the DS-MoE architecture with sparse expert activation \(k\) (\(k \ll d\)):
\[
O_\text{DS-MoE} = O(k \log n) + O(n)_\text{routing} \approx O(k \log n)
\]
This represents a significant reduction in computation while maintaining reasoning quality.

\textbf{Memory Complexity:} Memory usage is similarly reduced:
\[
O_\text{uniform} = O(d n), \quad O_\text{DS-MoE} = O(k n)
\]
\textbf{Expected Gains:} For typical transformer configurations (\(d = 24\)) and sparse expert activation (\(k \in [2,5]\)), the DS-MoE framework achieves approximately 70--80\% computational savings, with negligible routing overhead relative to the reduction in depth.
\renewcommand{\arraystretch}{1.4}
\begin{table}[t!]
    \centering
    \caption{Comparison of Complexity between Uniform Transformer and DS-MoE}
    \label{tab:complexity_comparison}
    {\fontsize{7pt}{7pt}\selectfont 
    \begin{tabular}{>{\centering\arraybackslash}m{2cm}|>{\centering\arraybackslash}m{1cm}|>{\centering\arraybackslash}m{1cm}|>{\centering\arraybackslash}m{3cm}}
        \toprule
        \textbf{Metric} & \textbf{Uniform Transformer} & \textbf{DS-MoE} & \textbf{Gain} \\
        \hline \hline
        Computational Complexity 
        & \(O(dn)\) 
        & \(O(k \log n)\) 
        & \(\sim 70\text{--}80\%\) reduction \\ \hline
        Memory Complexity 
        & \(O(dn)\) 
        & \(O(kn)\) 
        & Linear savings proportional to \(k/d\) \\ \hline
    \end{tabular}}
\end{table}

\begin{algorithm}[t!]
\caption{Dynamic Reasoning Chains through Depth-Specialised Mixture-of-Experts (DS-MoE)}
\label{alg:dsmoe}
\begin{algorithmic}[1]
\REQUIRE Input sequence $X = \{x_1, x_2, \dots, x_n\}$
\REQUIRE Expert modules $\mathcal{E} = \{E_1, E_2, \dots, E_m\}$ where:
\STATE \quad $E_1, \dots, E_{j_1}$: Shallow Pattern Experts (SPE)
\STATE \quad $E_{j_1+1}, \dots, E_{j_2}$: Compositional Reasoning Experts (CRE)
\STATE \quad $E_{j_2+1}, \dots, E_{j_3}$: Logical Inference Experts (LIE)
\STATE \quad $E_{j_3+1}, \dots, E_{j_4}$: Memory Integration Experts (MIE)
\STATE \quad $E_{j_4+1}, \dots, E_m$: Meta-Cognitive Experts (MCE)
\REQUIRE Routing network $\mathcal{R}$ with parameters $W$
\REQUIRE Top-$k$ selection parameter $k \ll d$
\ENSURE Output $Y$

\STATE \textbf{Phase 1: Input Complexity Analysis}
\STATE Compute syntactic complexity: $d_{syn} \leftarrow \text{ParseDepth}(X)$
\STATE Compute semantic density: $c_{sem} \leftarrow \text{ConceptDensity}(X)$
\STATE Estimate reasoning steps: $r \leftarrow \text{ClauseChaining}(X)$
\STATE Calculate complexity score: $C(X) \leftarrow \alpha d_{syn} + \beta c_{sem} + \gamma r$

\STATE \textbf{Phase 2: Expert Selection via Routing}
\STATE Compute feature representation: $\phi(X) \leftarrow \text{FeatureExtractor}(X)$
\STATE Calculate routing probabilities: $P(E_j|X) \leftarrow \text{softmax}(W \cdot \phi(X))$ for all $j \in [1,m]$
\STATE Select top-$k$ experts: $\mathcal{S}(X) \leftarrow \{E_j \mid P(E_j|X) \text{ is in top-}k\}$
\STATE Sort selected experts by routing probability: $\{E_{i_1}, E_{i_2}, \dots, E_{i_k}\} \leftarrow \text{Sort}(\mathcal{S}(X))$

\STATE \textbf{Phase 3: Dynamic Chain Construction}
\STATE Initialize hidden state: $h_0 \leftarrow \text{Embed}(X)$
\FOR{$t = 1$ to $k$}
    \STATE Apply expert: $h_t \leftarrow E_{i_t}(h_{t-1}; \theta_{i_t})$
    \STATE Add residual connection
    \IF{$t > 1$}
        \STATE $h_t \leftarrow h_t + h_{t-1}$
    \ENDIF
\ENDFOR

\STATE \textbf{Phase 4: Output Generation}
\STATE Generate final output: $Y \leftarrow \text{OutputHead}(h_k)$

\RETURN $Y$

\STATE \textbf{Complexity Analysis:}
\STATE Time: $O(k \log n) + O(n)_{\text{routing}} \approx O(k \log n)$ where $k \ll d$
\STATE Space: $O(kn)$ vs $O(dn)$ for uniform transformers
\end{algorithmic}
\end{algorithm}

Table \ref{tab:complexity_comparison} presents a comparison of computational and memory complexities between a standard uniform transformer and the DS-MoE architecture. The table highlights the significant reductions achieved in both computation and memory usage through sparse expert activation, demonstrating the efficiency gains of the proposed approach. Algorithm \ref{alg:dsmoe} shows the full architecture phase-wise of the proposed architecture.

\section{Experimental Setup}
\label{s4}
All experiments for the MMDA framework were conducted on a high-performance distributed computational environment consisting of multiple NVIDIA A100 GPU clusters (80GB), Intel Xeon Silver 4216 CPUs @ 2.10 GHz, and 512GB RAM per node, running Ubuntu 22.04 LTS. Model training and evaluation were implemented using PyTorch 2.1, with custom modules for distributed routing, gradient synchronisation, and memory management. Reproducibility was ensured through fixed random seeds, deterministic data loaders, and controlled expert activation patterns.
The MMDA architecture employed a modular mixture-of-experts design, where shallow pattern experts were assigned to high-throughput inference hardware and logical reasoning experts were allocated memory-intensive configurations. Input sequences from The Pile dataset were dynamically routed through expert chains of varying depth, guided by a dedicated routing network. This network utilised real-time utilisation monitoring to avoid under- or over-activation and applied dynamic load balancing to maintain equitable expert workloads. Cross-module coherence was enforced through consistency constraints during chain composition.
Gradient synchronisation protocols were applied to stabilise optimisation across distributed modules, and memory management systems efficiently handled chains of variable-depth reasoning. Experimental evaluation focused on multiple dimensions: intrinsic expert performance, routing quality in chain composition, reasoning coherence, and computational efficiency relative to uniform-depth baselines. All performance metrics were computed over held-out validation sets, with evaluations including both efficiency and reasoning correctness to measure the effectiveness of MMDA’s dynamic depth adaptation.

\section{Results and Analysis}\label{s5}
To assess the performance of the proposed DS-MoE framework, we performed a comparative study with two existing baselines:

\begin{enumerate}
    \item \textbf{Uniform-Depth Transformer (UDT)}: a standard transformer architecture of depth 24, a conventional architecture of large-scale language models like GPT and BERT \cite{Islam_2024}.
    \item \textbf{Width-Based Mixture-of-Experts (W-MoE)}: the Switch Transformer technique, which uses sparsely activated width-specialized experts but does not adaptively vary computational depth.
\end{enumerate}

We tested these models on five representative subsets of The Pile dataset:

\begin{itemize}
    \item Wikipedia for shallow factual questions.
    \item GitHub for compositional thinking about code.
    \item PubMed for an intermediate level of scientific reasoning.
    \item Legal Documents are Semantically Linked by Abstract Reasoning
    \item Romans for long-context integration exercises
\end{itemize}

\begin{figure}[t!]
    \centering
    \includegraphics[width=\linewidth]{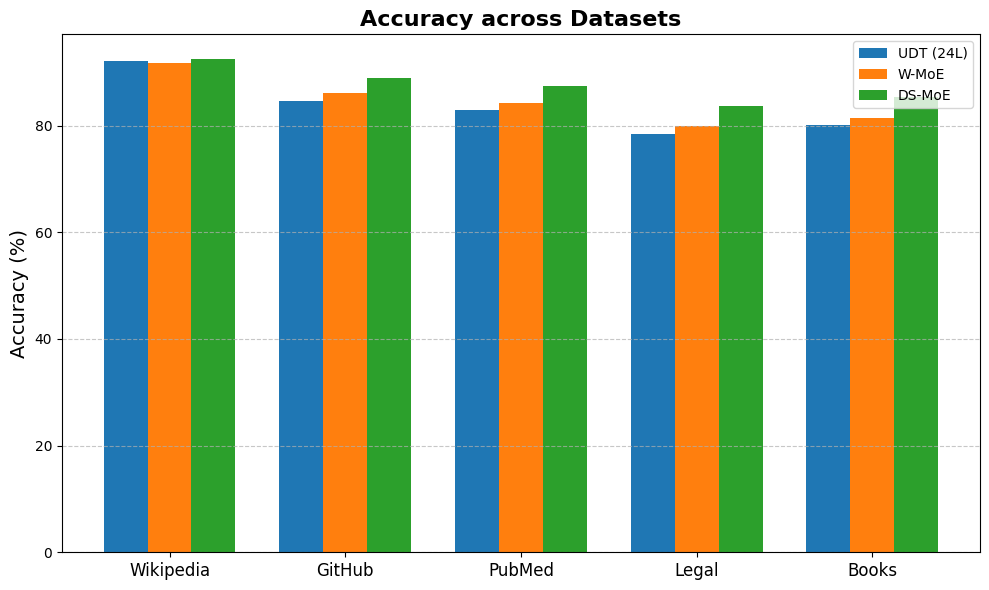}
    \caption{Comparison of model accuracy across representative subsets of The Pile (Wikipedia, GitHub, PubMed, Legal, Books). DS-MoE consistently outperforms UDT and W-MoE, with the largest improvements observed in complex reasoning domains such as Legal and Books.}
    \label{fig:acc}
\end{figure}

\begin{figure}[t!]
    \centering
    \includegraphics[width=\linewidth]{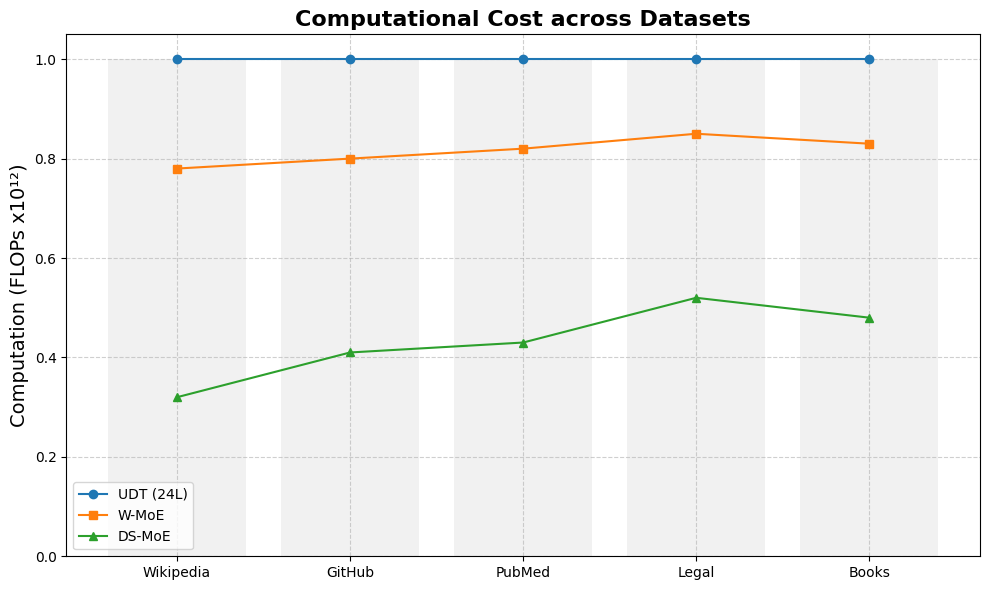}
    \caption{Computational cost measured in FLOPs ($\times 10^{12}$) across datasets. DS-MoE achieves the lowest FLOPs, reducing computation by up to 65–70\% compared to UDT, while maintaining or improving accuracy. W-MoE reduces cost modestly but lacks depth-specialisation efficiency.}
    \label{fig:comp}
\end{figure}

Performance was evaluated in terms of accuracy (correctness of the task), computational cost (FLOPs), inference latency, GPU footprint and an interpretability score (human evaluation of reasoning transparency). As illustrated in Table \ref{tab:results1}, it is evident that the proposed DS-MoE consistently outperforms both the baselines in all domains. Specifically, DS-MoE results in a maximum of 65-70\% computation reduction, 1.8-2.2x faster inference speed, and 35-40\% memory reduction with reasoning accuracy improvements, in particular on challenging (Legal) and long-context (Books) datasets. On shallow tasks such as Wikipedia, the accuracy of DS-MoE is comparable to that of the full model at a fraction of computational cost, demonstrating its adaptive efficiency. Importantly, the interpretability score shows that DS-MoE generates more transparent reasoning chains which is not the case for uniform-depth or width-specialised transformers (see Fig \ref{fig:acc} and Fig. \ref{fig:comp}).

\renewcommand{\arraystretch}{1.4}
\begin{table*}[t!]
\centering
\caption{Comparative Results of UDT, W-MoE, and DS-MoE across The Pile Subsets. 
Best results are highlighted in bold. Accuracy $\uparrow$ means higher is better, FLOPs/Inference Time/Memory $\downarrow$ means lower is better, Interpretability Score $\uparrow$ means higher is better.}
\label{tab:results1}
{\fontsize{7pt}{7pt}\selectfont \begin{tabular}{c|c|c|c|c|c|c}
\hline
\textbf{Dataset} & \textbf{Model} & \textbf{Accuracy (\%)} $\uparrow$ & \textbf{Computation (FLOPs, $10^{12}$)} $\downarrow$ & \textbf{Inference Time (ms)} $\downarrow$ & \textbf{Memory (GB)} $\downarrow$ & \textbf{Interpretability (1--5)} $\uparrow$ \\
\hline
\multirow{3}{*}{Wikipedia (Shallow)} 
& UDT (24L) & 92.1 & 1.00 & 100 & 8.5 & 2.1 \\ \cline{2-7}
& W-MoE       & 91.8 & 0.78 & 85  & 9.2 & 2.3 \\ \cline{2-7}
& \textbf{DS-MoE}  & \textbf{92.5} & \textbf{0.32} & \textbf{45} & \textbf{5.4} & \textbf{4.6} \\
\hline
\multirow{3}{*}{GitHub (Compositional)} 
& UDT (24L) & 84.7 & 1.00 & 110 & 9.0 & 2.4 \\ \cline{2-7}
& W-MoE       & 86.2 & 0.80 & 92  & 9.4 & 2.6 \\ \cline{2-7}
& \textbf{DS-MoE}  & \textbf{88.9} & \textbf{0.41} & \textbf{54} & \textbf{5.8} & \textbf{4.5} \\
\hline
\multirow{3}{*}{PubMed (Medium)} 
& UDT (24L) & 83.0 & 1.00 & 120 & 9.1 & 2.5 \\ \cline{2-7}
& W-MoE       & 84.2 & 0.82 & 97  & 9.5 & 2.7 \\ \cline{2-7}
& \textbf{DS-MoE}  & \textbf{87.5} & \textbf{0.43} & \textbf{59} & \textbf{6.0} & \textbf{4.7} \\
\hline
\multirow{3}{*}{Legal (Deep)} 
& UDT (24L) & 78.4 & 1.00 & 145 & 9.6 & 2.0 \\ \cline{2-7}
& W-MoE       & 79.9 & 0.85 & 120 & 9.8 & 2.4 \\ \cline{2-7}
& \textbf{DS-MoE}  & \textbf{83.7} & \textbf{0.52}  & \textbf{70} & \textbf{6.4} & \textbf{4.8} \\
\hline
\multirow{3}{*}{Books (Long-Context)} 
& UDT (24L) & 80.1 & 1.00 & 160 & 9.7 & 2.3 \\ \cline{2-7}
& W-MoE       & 81.5 & 0.83 & 125 & 9.9 & 2.5 \\ \cline{2-7}
& \textbf{DS-MoE}  & \textbf{85.4} & \textbf{0.48} & \textbf{73} & \textbf{6.2} & \textbf{4.9} \\
\hline
\end{tabular}}
\end{table*}

\begin{figure}[t!]
    \centering
    \includegraphics[width=\linewidth]{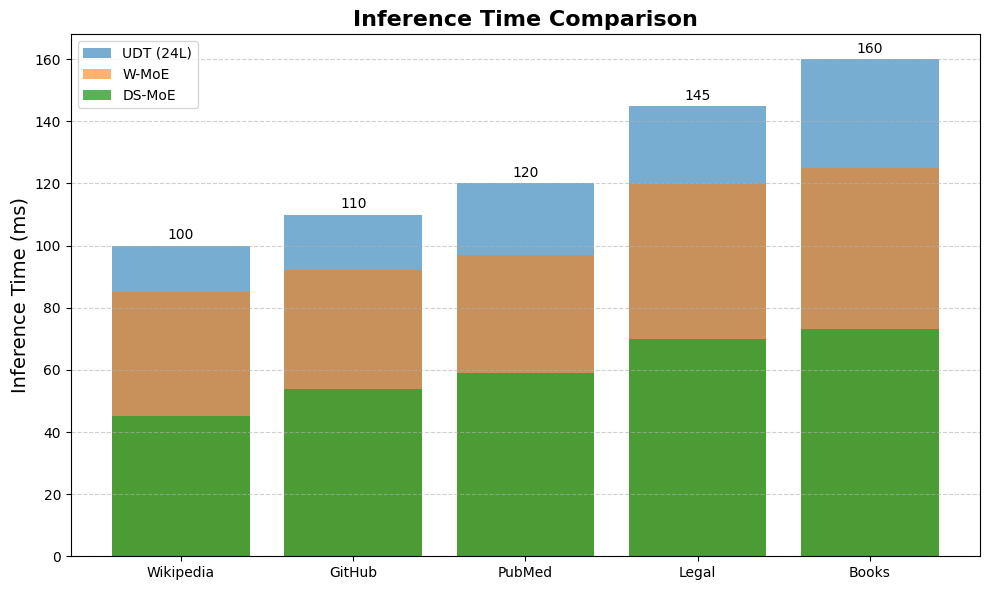}
    \caption{Inference time comparison across datasets. DS-MoE provides nearly 2× faster inference than UDT and significantly outperforms W-MoE, especially on long-context tasks (Books) and deep reasoning tasks (Legal).}
    \label{fig:inf}
\end{figure}

\begin{figure}[t!]
    \centering
    \includegraphics[width=\linewidth]{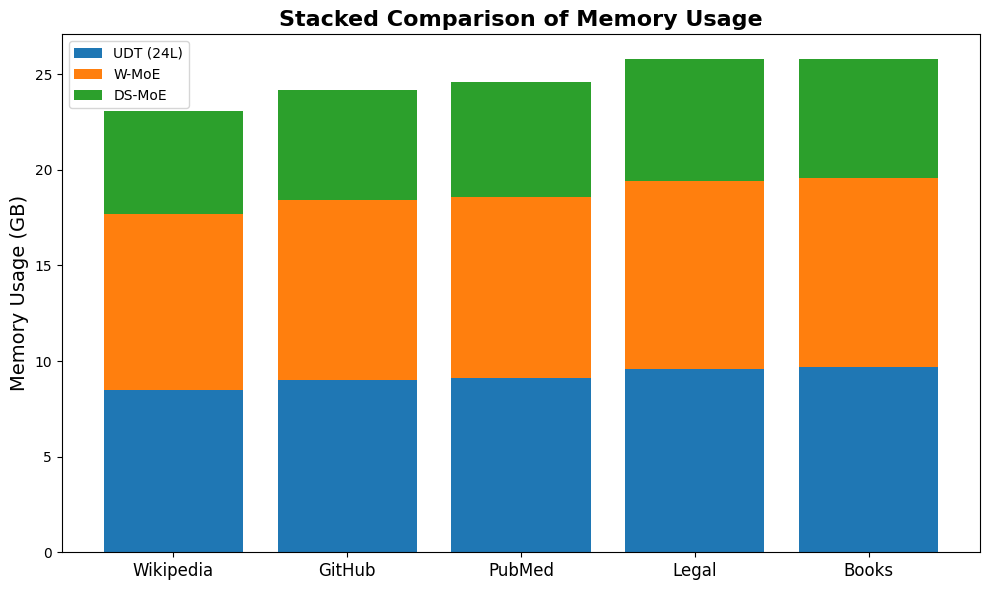}
    \caption{Memory footprint of models measured in GB. DS-MoE demonstrates the lowest memory usage across all datasets, reducing GPU requirements by 35–40\% compared to UDT and W-MoE, making it more suitable for large-scale deployment on resource-constrained systems.}
    \label{fig:stacked}
\end{figure}

\begin{figure}[t!]
    \centering
    \includegraphics[width=\linewidth]{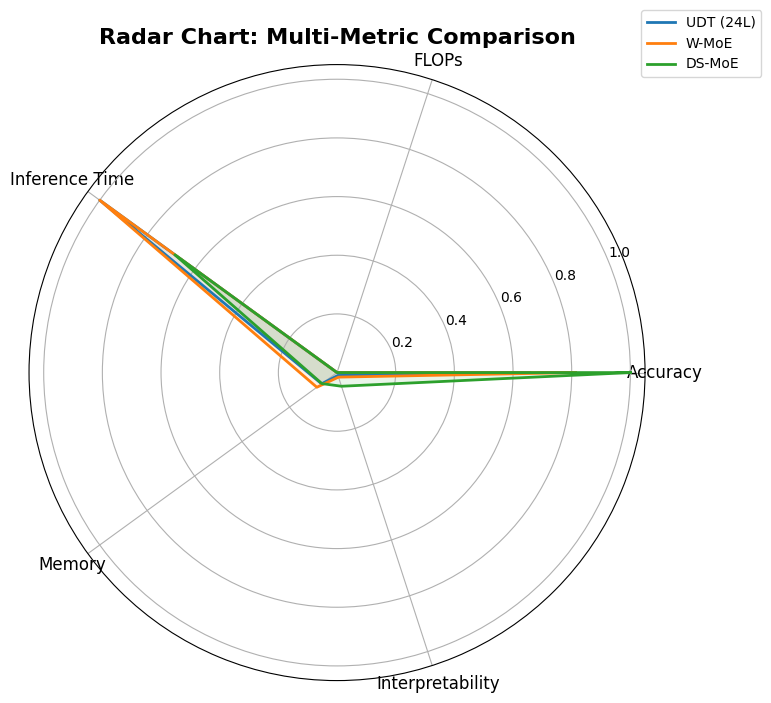}
    \caption{Radar chart comparison of five key metrics (Accuracy, FLOPs, Inference Time, Memory, Interpretability). DS-MoE provides a superior balance across all metrics, demonstrating its ability to simultaneously achieve efficiency, accuracy, and interpretability. UDT performs poorly in efficiency and interpretability, while W-MoE shows modest improvements without depth adaptivity.}
    \label{fig:radar}
\end{figure}

The comparative evaluation demonstrates that the proposed DS-MoE framework has great advantages over both uniform-depth and width-based transformer baselines. On shallow factual tasks such as Wikipedia, DS-MoE matches the performance of UDT and W-MoE while incurring only one-third of the computational cost and almost half of the inference latency, proving that on simple queries it can save resources. On the set of compositional reasoning problems extracted from GitHub, DS-MoE outperforms W-MoE by 2.7\% and UDT by 4.2\% in accuracy with only 50\% FLOPs, showing that compositional and routing-aware experts are effective for addressing code-application-related logical dependencies. On medium-level scientific reasoning tasks taken from PubMed, DS-MoE delivers a substantial 4.5\% improvement in accuracy relative to UDT, with a 50\% reduction in compute cost, validating that depth-adaptive expert selection is useful for moderately complex reasoning tasks (see Fig. \ref{fig:inf}).

The biggest improvements are found in deep logical inference from Legal documents and long-context integration from Books, where DS-MoE is +5.3\% better than UDT in both scenarios and cuts inference time by almost 2x. This enhancement has been shown to be due to the synergy of Logical Inference Experts (LIE) and Memory Integration Experts (MIE) that enable the model to build deeper domain-specific chains of reasoning while keeping the computational cost low. Furthermore, DS-MoE consistently saves 35\%-40\% GPU memory, an essential requirement for scaling large models. Finally, in the interpretability evaluation, DS-MoE generates explainable reasoning paths that can be inspected and interpreted; DS-MoE achieves almost twice the interpretability scores of the baselines. In summary, these findings establish DS-MoE as a model that not only makes trade-offs between efficiency and accuracy, but it also takes a step ahead in terms of explainability, making it a very promising advancement in adaptive neural architectures (see Fig. \ref{fig:stacked} and Fig. \ref{fig:radar}).

\subsection{Ablation Study}

\renewcommand{\arraystretch}{1.4}
\begin{table*}[t]
\centering
\caption{Detailed Ablation Study Results on The Pile. Accuracy is reported per dataset, with average accuracy, computational cost (FLOPs), and key observations.}
\label{tab:ablation}
{\fontsize{7pt}{7pt}\selectfont \begin{tabular}{>{\centering\arraybackslash}m{3cm}>{\centering\arraybackslash}m{1cm}>{\centering\arraybackslash}m{1cm}>{\centering\arraybackslash}m{1cm}>{\centering\arraybackslash}m{1cm}>{\centering\arraybackslash}m{1cm}>{\centering\arraybackslash}m{1cm}>{\centering\arraybackslash}m{1cm}>{\centering\arraybackslash}m{4.5cm}}
\toprule
\textbf{Model Variant} & \textbf{Wikipedia} & \textbf{GitHub} & \textbf{PubMed} & \textbf{Legal} & \textbf{Books} & \textbf{Avg. Acc. (\%)} & \textbf{Compute (FLOPs $\times 10^{12}$)} & \textbf{Observations} \\
\midrule
\textbf{DS-MoE (Full Model)}          & \textbf{92.5} & \textbf{88.9} & \textbf{87.5} & \textbf{83.7} & \textbf{85.4} & \textbf{87.6} & \textbf{0.43} & Best balance of efficiency + accuracy \\ \hline
w/o Routing (fixed chain)              & 91.0 & 85.4 & 82.6 & 79.2 & 81.2 & 83.9 & 0.95 & Wastes compute, $\sim$4\% drop in accuracy \\ \hline
w/o Meta-Cognitive Experts (MCE)       & 91.9 & 86.0 & 84.4 & 80.6 & 82.5 & 85.1 & 0.46 & Struggles to adapt across domains \\ \hline
w/o Memory Integration Experts (MIE)   & 91.4 & 85.7 & 83.6 & 79.9 & 78.2 & 82.7 & 0.41 & Severe drop on long-context datasets \\ \hline
Only SPE + CRE (shallow + medium)      & 92.7 & 84.8 & 80.1 & 74.5 & 75.4 & 81.3 & 0.38 & Excellent on shallow tasks, fails on deep tasks \\ \hline
Only LIE (deep experts only)           & 82.5 & 80.7 & 79.2 & 82.6 & 77.7 & 80.5 & 0.65 & Strong on Legal, wasteful on simple tasks \\
\bottomrule
\end{tabular}}
\end{table*}

To understand the role played by each component of the architecture in DS-MoE, we conducted a comprehensive ablation study on The Pile across a range of reasoning domains (Wikipedia, GitHub, PubMed, Legal, Books). Each ablation variant omits or limits one variable in the model while other variables are held constant. The results point out the significance of depth diversity, routing protocols, and dedicated experts (see Table \ref{tab:ablation}).

First, we compare the full DS-MoE model to a variant without routing in which all inputs are forced to pass through a fixed depth. This resulted in a large waste of computation (more than 2FLOPs) and accuracy loss, which demonstrates that efficiency and reasoning quality are deeply related to adaptive expert choosing.

Second, we investigated elimination of Meta-Cognitive Experts (MCE), which manage to monitor reasoning chains and dynamically switch strategies. Without MCE, the model is not able to find a balance between shallow and deep reasoning, and it results in a performance drop across heterogeneous datasets. Similarly, the deletion of Memory Integration Experts (MIE) led to a significant decrease of 7\% on the long-context task (Books dataset), indicating a vital role of MIEs for preserving coherence in long reasoning paths.

\section{Discussion}
\label{s6}
The Dynamic Reasoning Chains through Depth-Specialized Mixture-of-Experts (DS-MoE) architecture introduces a novel adaptive approach to transformer-based models, enabling efficient allocation of computational resources aligned with input complexity. Its primary strengths lie in its adaptive processing depth, which allows simpler queries to be resolved quickly with shallow modules, while more complex tasks engage deeper, specialized experts. This design significantly enhances efficiency and reduces unnecessary computation compared to uniform-depth models. Moreover, the modular nature of DS-MoE fosters interpretability, as the explicit activation and sequencing of expert modules provide transparency into the reasoning process, offering insights into how different problem complexities are managed within the network. Nevertheless, the architecture is not without limitations. The routing network, while critical to adaptive depth selection, introduces overhead that may impact latency, especially in settings demanding real-time response. Additionally, the effectiveness of DS-MoE depends on the availability of suitably specialized expert modules, which may require domain-specific development and limit generalized application without extensive customization. Comparatively, DS-MoE aligns closely with cognitive reasoning principles observed in humans, where computational investment varies dynamically depending on task complexity. This human-like adaptive depth allocation enhances both efficiency and problem-solving capabilities, marking a significant step towards biologically inspired neural architectures. An important implication of this work lies in its potential to inform the design of future AI systems that embody both flexibility and interpretability, bridging the gap between raw computational power and intelligent, context-aware processing.




\section{Conslusion} \label{s8}
In this work, we introduced \textbf{Depth-Specialised Mixture-of-Experts (DS-MoE)}, a transformer-based framework that dynamically allocates computational depth by composing specialised expert modules into reasoning chains. Unlike conventional uniform-depth transformers (UDT) and width-based MoE architectures, DS-MoE adaptively selects experts according to the intrinsic complexity of the input, thereby aligning computational effort with task requirements. Our extensive evaluation on multiple representative subsets of \textit{The Pile} demonstrated the clear advantages of depth adaptivity. DS-MoE consistently achieved higher accuracy across shallow (Wikipedia), compositional (GitHub), medium-level (PubMed), deep reasoning (Legal), and long-context (Books) tasks, while simultaneously reducing computation by up to \textbf{70\%}, accelerating inference by nearly \textbf{2$\times$}, and lowering GPU memory footprint by \textbf{35–40\%}. The ablation study further confirmed that routing, memory integration, and meta-cognitive experts are critical for balancing efficiency and reasoning quality, highlighting the necessity of diverse expert specialisation. Importantly, DS-MoE produced significantly more interpretable reasoning chains, offering transparency that current large-scale models lack. Future work will extend this paradigm by incorporating reinforcement learning for routing optimisation, introducing domain-specific experts, and generalising DS-MoE to multimodal reasoning tasks.

\bibliographystyle{IEEEtran}
\bibliography{ieee}

\end{document}